\documentclass[11pt]{article}

\usepackage{acl}

\usepackage{times}
\usepackage{latexsym}

\usepackage[T1]{fontenc}

\usepackage[utf8]{inputenc}

\usepackage{microtype}

\usepackage{inconsolata}

\usepackage{amsmath}
\usepackage{amssymb}
\usepackage{subcaption}
\usepackage{graphicx}
\usepackage{booktabs}
\usepackage{multirow}
\usepackage{soul}
\usepackage{float}
\usepackage{colortbl,xcolor}
\usepackage{algorithm}
\usepackage{algpseudocode}
\usepackage{algorithmicx}
\usepackage{rotating}

\title{Pruning Large Language Models to Intra-module Low-rank Architecture with Transitional Activations}

\author{
  Bowen Shen$^{1,2}$\footnotemark[2]\ , Zheng Lin$^{1,2~*}$\ , Daren Zha$^{1~*}$ , Wei Liu$^{3}$, Jian Luan$^{3}$, Bin Wang$^{3}$,\\
  \bf Weiping Wang$^{1}$\\
  $^{1}$Institute of Information Engineering, Chinese Academy of Sciences, Beijing, China\\
  $^{2}$School of Cyber Security, University of Chinese Academy of Sciences, Beijing, China\\
  $^{3}$Xiaomi AI Lab, Beijing, China\\
  \texttt{\{shenbowen, linzheng, zhadaren, wangweiping\}@iie.ac.cn}\\
  \texttt{\{liuwei40, luanjian, wangbin11\}@xiaomi.com}
}
\begin{document}
\maketitle
\begin{abstract}
  Structured pruning fundamentally reduces computational and memory overheads of large language models (LLMs) and offers a feasible solution for end-side LLM deployment.
Structurally pruned models remain dense and high-precision, highly compatible with further tuning and compression.
However, as the coarse-grained structured pruning poses large damage to the highly interconnected model, achieving a high compression ratio for scaled-up LLMs remains a challenge.
In this paper, we introduce a task-agnostic structured pruning approach coupled with a compact Transformer architecture design. The proposed approach, named TransAct, reduces \ul{trans}itional \ul{act}ivations inside multi-head attention (MHA) and multi-layer perceptron (MLP) modules, while preserving the inter-module activations that are sensitive to perturbations. Hence, the LLM is pruned into an intra-module low-rank architecture, significantly reducing weights, KV Cache and attention computation.
TransAct is implemented on the LLaMA model and evaluated on downstream benchmarks. Results verify the optimality of our approach at high compression with respect to both efficiency and performance. Further, ablation studies reveal the strength of activation-guided iterative pruning and provide experimental analysis on the redundancy of MHA and MLP modules.

  \renewcommand{\thefootnote}{\fnsymbol{footnote}}
  \footnotetext[2]{\ \ Work done during an internship at Xiaomi AI Lab.}
  \footnotetext[1]{\ \ Zheng Lin and Daren Zha are the corresponding authors.}
\end{abstract}

\section{Introduction}

Deploying large language models (LLMs) locally on edge devices instead of relying on remote APIs has been a pressing initiative. Local deployment of LLMs ensures independence from network conditions and enhances privacy at an advanced level \cite{ma2023pipellm}.
Nevertheless, deploying a scaled-up LLM onto a resource-constrained end device poses multifaceted challenges, encompassing inference speed, memory footprint, and power consumption.
Therefore, comprehensive optimizations on the efficiency of LLMs are imperative, including architecture design \cite{gu2023mamba}, model compression \cite{zhu2023survey}, inference schemes \cite{leviathan2023fast, cai2024medusa}, compilation and runtime \cite{lai2023relax}.

\begin{figure}[t]
    \centering
    \includegraphics[width=0.9\columnwidth]{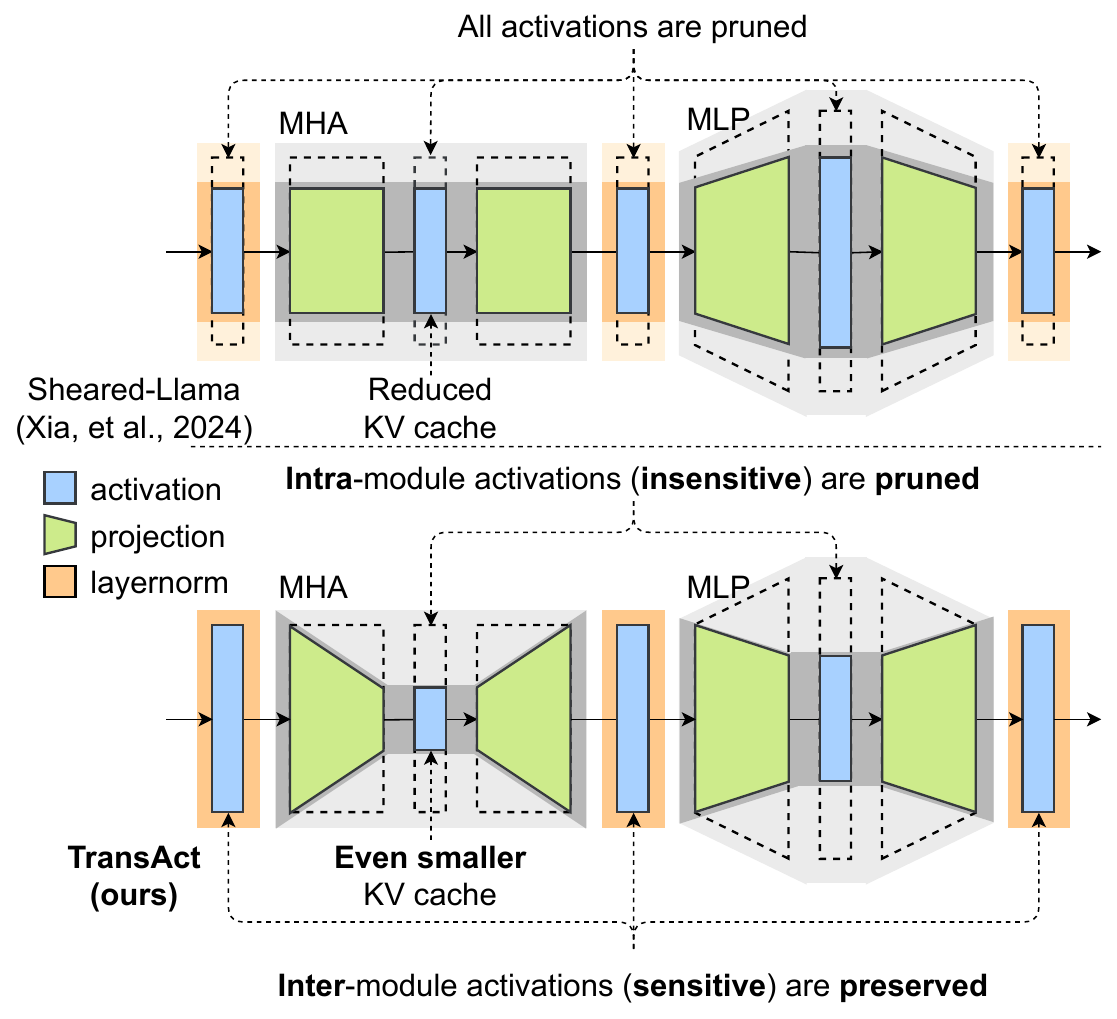}
    \caption{An illustration of TransAct model architecture. The model weights and activations are colored green and blue, respectively. Dashed hollow blocks represent the weights and activations that are pruned out.}
    \label{fig:intro}
\end{figure}

Model compression emerges as the silver-bullet solution for reducing deployment costs given an accessible LLM.
To essentially reduce model computation and memory overhead, pruning aims to discard weights with low salience to the LLM.
\citet{jaiswal2023compressing} suggest that state-of-the-art (SOTA) unstructured pruning approaches i.e., SparseGPT \cite{frantar2023sparsegpt} and Wanda \cite{sun2023simple}, along with their semi-structured variations, often underperform in downstream benchmarks. \citet{zimmer2023perp} emphasize the significance of post-training after pruning to restore the capabilities of the LLM.
However, the post-training and inference of a sparse model are notably inefficient.
Also, an unstructured pruning with arbitrary sparsity pattern has no speedup or memory saving on the LLM, whereas a semi-structured sparse model heavily relies on specific hardware \cite{frantar2023sparsegpt}.

An alternative pruning category, i.e., structured pruning, has shown feasibility for LLMs.
LLM-Pruner \cite{ma2023llm}, the pioneering structured pruning of LLM, incorporates the approximated Taylor series as the pruning metric. However, this approximation loses accuracy when pruning a large ratio of the model \cite{NIPS1989_6c9882bb}. While Taylor expansion assumes small perturbations, it is not applicable when a large number of parameters are pruned (i.e., set to zero).
The SOTA approach Sheared-LLaMA \cite{xia2024sheared}, on the other hand, completely transfers the evaluation of the pruning metric to supervised training with masks. However, training with masks poses much more computation and memory footprint at training time, as well as the training unstableness.
Also, the pruned architecture of Sheared-LLaMA, as illustrated in the upper part of \autoref{fig:intro}, involves the unified pruning of layer normalization (LN) weights, disregarding the varying sensitivity of LN parameters to perturbation across layers \cite{zhao2023unveiling}.

To address the challenges of efficient and effective LLM pruning, we propose TransAct, a transitional activation-based structured pruning approach.
From the perspective of pruning architectural design, TransAct reduces intra-module activations, which prunes the MHA and MLP in LLM into low intrinsic dimension as depicted in \autoref{fig:intro}.
TransAct pruning metric is inspired by the observation of \citet{dettmers2022llm} that a small proportion of activations within the LLM exhibit outlier magnitudes, rendering them particularly sensitive to perturbations and need to be preserved.
This approach effectively reduces the memory footprint of both model weights and KV cache, alleviating the memory constraints inherent in autoregressive generation on edge devices \cite{kwon2023efficient}.
Specifically, the contributions of this paper are outlined as follows.

\begin{itemize}
    \item We propose a co-design of pruning architecture and pruning metric named TransAct, which substantially compresses the KV cache as well as the model weights.
    \item TransAct pruning architecture achieves the fastest inference speed among SOTA pruned models, while the pruning is efficient without gradients or masked training.
    \item Experiment results on downstream benchmarks verified the stableness of TransAct at a high compression ratio. Ablation studies on module redundancy provide insights for compact model design.
\end{itemize}

\section{Related Work}\label{sec:related}

Extensive works have been proposed to optimize the efficiency of Transformer-based LMs, covering pruning, quantization, dynamic acceleration, etc. However, to generalize these approaches to the continually scaling-up LLMs remains challenging.

Quantization, which reduces the bit representation of values, stands out due to its ease of implementation.
Post-training quantization (PTQ) approaches, e.g., GPTQ \cite{frantar2022gptq} and AWQ \cite{lin2023awq}, are without any further tuning after the quantization. On the contrary, quantization-aware training (QAT) approaches train the model along with the quantization parameters and is still challenging when the LLM is scaled up \cite{liu2023llm}.
Quantizing an LLM from \texttt{float16} to \texttt{int3} with weight-only PTQ approaches like GPTQ \cite{frantar2022gptq} can reach roughly 80\% compression of model weights. However, the KV cache which contributes to a large amount of memory overheads is still in \texttt{float16} and uncompressed. Also, obtaining an acceptable quantization precision with \texttt{int3} weights remains a challenge. \citet{xiao2023smoothquant} proposed a W8A8 PTQ approach where both weights and activations are quantized to \texttt{int8}, saving 50\% memory footprint.
The lack of flexibility poses a significant limitation to quantization. Most general computing platforms and libraries primarily support low-bit representations such as \texttt{int8} and \texttt{int4} \cite{nagel2021white}. However, opting for representations lower than 4 bits necessitates dequantization back to the supported higher-bit representations, thereby introducing additional computation and memory overheads.

\begin{figure*}[t]
    \centering
    \includegraphics[width=0.95\textwidth]{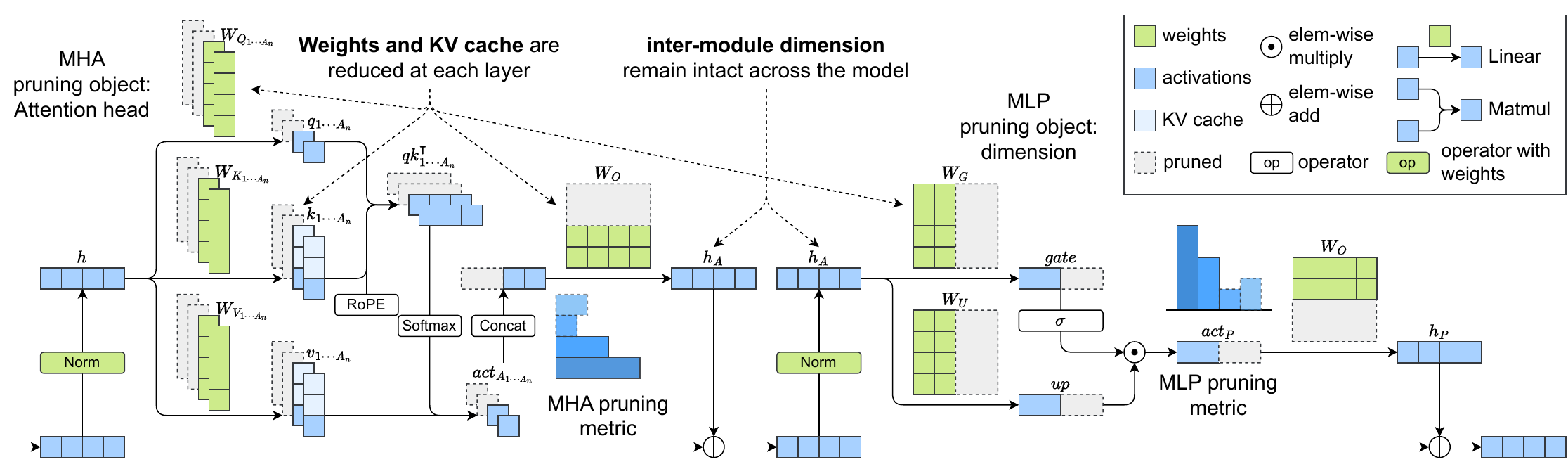}
    \caption{Detailed TransAct workflow on a Transformer layer with MHA and gated MLP. Bar charts indicate the activation-based pruning metric.}
    \label{fig:arch}
\end{figure*}

Apart from quantization, unstructured pruning is also an efficient approach to obtain a sparse LLM. \citet{frantar2023sparsegpt} and \citet{sun2023simple} enabled fully unstructured and semi-structured N:M sparsity (i.e., N zeros in M consecutive weights) of LLM across different sizes. However, there are two major obstacles hindering the adoption of unstructured sparsity. (1) The pruned sparse LLM cannot be efficiently further trained. Although \citet{sun2023simple} claimed to use LoRA \cite{hu2021lora} to train the compressed model, the LoRA modules cannot be merged into the sparse backbone LLM, which incurs additional overhead at inference time.
(2) The sparsity is fixed at 50\% with current hardware and platform affinity. While only NVIDIA Ampere and Hopper GPUs support the 2:4 sparsity pattern, achieving customized sparsity requires hardware co-design \cite{Fang_2022}. This limitation restricts the broader application of unstructured pruning.

The approaches discussed above are static compression of LLM, where the computation at inference is fixed. On the contrary, dynamic acceleration at inference time speeds up LLM generation by selective computation. Early exiting approaches \cite{schuster2022confident,delcorro2023skipdecode} allow the LLM to finish the decoding of a token without passing all the layers. Mixture-of-Expert (MoE) architecture \cite{jiang2024mixtral, lepikhin2020gshard} incorporates multiple parameter shards in MLP as experts and selects experts to compute when facing different inputs. The dynamic approaches usually do not reduce parameters. Thus, the storage of the model is not reduced, while the runtime memory can be saved by fine-grained neuron-aware offloading \cite{song2023powerinfer}.

\section{Methodology}\label{sec:methodology}

In this section, we first recap the preliminaries of Transformer-based LLM architecture and introduce the transitional activations. Then, we propose our approach TransAct with the pruning metric and architecture design of the pruned model.

\subsection{Transitional Activations in LLM}\label{sec:preliminaries}

Transformer-based LLMs generally consist of embedding, MHA (multi-head attention), MLP (multi-layer perceptron), and LM head.

The majority of model weights lie in MHA and MLP, which exist in every Transformer layer of the LLM. Specifically, MHA has three matrices $\boldsymbol{W_Q}$, $\boldsymbol{W_K}$, $\boldsymbol{W_V}$ with the shape of $H \times A$, and one matrix $\boldsymbol{W_O}$ of the inverted. The MHA mechanism splits the output dimension $A$ into $A_n \times A_d$ (i.e., head number by head dimension), which forms $A_n$ logical attention heads. The input activation $\boldsymbol{h}^l$ of the $l$-th layer is projected by $\left\{\boldsymbol{W_Q}^l_k, \boldsymbol{W_K}^l_k, \boldsymbol{W_V}^l_k\right\}_{k=1}^{A_n}$ and split into $A_n$ groups of query, key, and value $\{\boldsymbol{q}^l_k, \boldsymbol{k}^l_k, \boldsymbol{v}^l_k\}_{k=1}^{A_n}$. Then the multi-head self-attention computation is as
\begin{gather}
    \boldsymbol{act_A}^l_k = \operatorname{Softmax}(\boldsymbol{q}^l_k {\boldsymbol{k}^l_k}^\intercal/\sqrt{A_d}) \boldsymbol{v}^l_k, \label{eq:mha}
\end{gather}
where $k$ is the attention head index counted from 1 to $A_n$, and $l$ indicates the $l$-th layer. ${H\times A_d}$ at the superscript is the shape annotation of the weight matrix. Then, the results are concatenated to shape $A$ and projected back to shape $H$ by $\boldsymbol{W_O}$.
\begin{gather}
    \boldsymbol{h_A}^l = \operatorname{Concat}[\boldsymbol{act_A}^l_k]_{k=1}^{A_n} {\boldsymbol{W_O}^l}^{A\times H}.
\end{gather}

As a bound between the group of $\boldsymbol{W_Q}$, $\boldsymbol{W_K}$, $\boldsymbol{W_V}$ and $\boldsymbol{W_O}$, we define $\boldsymbol{act_A}^l$ as the transitional activation of MHA module. By default, the transitional size $A$ of MHA is the same as hidden dimension $H$, but $A$ can be smaller than $H$ by reducing $A_n$ or $A_d$ in the case of pruning.

The other module, MLP, has a pair of upcast and downcast phases. In the first phase, the input hidden state $\boldsymbol{h}$ is projected to a transitional state with larger dimension $P$ through $\boldsymbol{W_U}$ and an optional gate $\boldsymbol{W_G}$, the later phase consists of a downcast $\boldsymbol{W_D}$ that projects the transitional state back to the original shape $H$. We consider $\boldsymbol{W_G}$ exists and formulate MLP as
\begin{gather}
    \boldsymbol{act_P}^l = \sigma(\boldsymbol{h_A}^l{\boldsymbol{W_G}^l}^{H\times P})\odot (\boldsymbol{h_A}^l{\boldsymbol{W_U}^l}^{H\times P}), \label{eq:mlp}\\
    \boldsymbol{h_P}^l = \boldsymbol{act_P}^l{\boldsymbol{W_D}^l}^{P\times H}.
\end{gather}
Similar to the MHA module, we define $\boldsymbol{act_P^l}$ as the transitional activation of the MLP module at the $l$-th layer. In case there is no optional gating in the model (e.g., OPT \cite{zhang2022opt}, BLOOM \cite{le2023bloom}), the transitional activation of the MLP module can be viewed as $\boldsymbol{act_P}^l = \sigma(\boldsymbol{h_A}^l{\boldsymbol{W_G}^l}^{H\times P})$.

\subsection{Pruning with Transitional Activations}\label{sec:object}

Based on the model architecture, we identify the pruning target as the following.
(1) $A_n$, i.e., the number of attention heads in MHA. On the other hand, $A_d$ is kept intact, as reducing it incurs the adaption of RoPE (rotary positional embedding) \cite{su2024roformer} used by a high quantity of LLMs and increases the training unstableness.
(2) $P$, i.e., the transitional dimension of MLP. Studies \cite{mirzadeh2023relu} indicate that, with an activation function suppressing negative values, the transitional state of MLP is with high redundancy.
It is worth mentioning that $H$, i.e., the hidden dimension throughout the model is not compressed. We justify the reason as compressing $H$ incurs the unified pruning of layer normalization (LN) weights across layers, whereas the sensitivity of LN parameters to perturbation is not unified across layers \cite{zhao2023unveiling}. Although further training can reconstruct the LN module from the damage of compressing, the significant training cost is contrary to efficiency.

With the definition of transitional activations and the pruning objects, we propose the transitional activation-based pruning approach to compress MHA and MLP modules into an intra-module low-rank architecture as depicted in \autoref{fig:arch}.
For the MHA module, we define the pruning granularity (i.e., the least separable structure) to be the attention head, in turn reducing $A_n$ while keeping $A_d$ intact. Such an attention head pruning is unified on $\boldsymbol{W_Q}$, $\boldsymbol{W_K}$ and $\boldsymbol{W_V}$ because the self-attention calculation, as formulated in \autoref{eq:mha}, requires the aligned head index among the three matrices.
Then, we can define the salience of all heads in MHA as
\begin{gather}
    {\mathcal{S}^l_\mathcal{A}}_k = \frac{1}{A_d}\sum\limits_{i=0}^{A_d}\left\lVert{\boldsymbol{act_A}^l_k}_i \right\rVert_2 +\alpha\max\limits_{i}\left\lVert{\boldsymbol{act_A}^l_k}_i \right\rVert_2, \label{eq:mha_score}
\end{gather}
where $\alpha$ is a weight factor amplifying the maximum activation value in the $k$-th head. By \autoref{eq:mha_score}, we want to evaluate both the general and outlier values in the activations, so that we can precisely prune out the most insignificant head.
For MLP, we can simply use the corresponding value of $\boldsymbol{act_P}$ to represent the salience of MLP transitional dimension as ${\mathcal{S}^l_\mathcal{P}}_i = \left\lVert\boldsymbol{act_P}^l_i \right\rVert_2$.

With the salience $\mathcal{S_A}$ and $\mathcal{S_P}$ formulated, we can model the activation-based structured pruning of a weight matrix $W$ as
\begin{gather}
    \operatorname{prune}(\boldsymbol{W},K,\mathcal{S})=\operatorname{Concat}[\boldsymbol{W}_i]_{i\in \arg\operatorname{topK}(\mathcal{S})}. \label{eq:prune}
\end{gather}
Specifically, the pruning dimension of $\boldsymbol{W_Q}$, $\boldsymbol{W_K}$, $\boldsymbol{W_V}$, $\boldsymbol{W_G}$ (optional) and $\boldsymbol{W_U}$ is the output, while the pruning dimension of $\boldsymbol{W_O}$ and $\boldsymbol{W_D}$ is the input as depicted in \autoref{fig:arch}.

Obtaining the salience of the source LLM requires only forward passes with a small amount of calibration samples. Hence, the pruning procedure is efficient in both memory and computation. To avoid a single shot pruning to compression ratio $R$ posing unrecoverable damage to the model, we provide an enhanced implementation where the model is iteratively pruned to the target size. A set of pruning ratios is defined as $\mathcal{R} = \{r_1, r_2, \cdots, r_n\}$, where the $i$-th shot prunes the model to the size of $(A^\prime_i, P^\prime_i)$ subject to $\sum_{r_i\in\mathcal{R}}=R$, and $A^\prime_i \mod A_d = 0$.
During the interval of two pruning steps, full fine-tuning (FT) is performed on the model to recover the pruning damage.

\section{Experimental Evaluation}

\subsection{Experiment Setup}

\paragraph{Model and Datasets Settings}

In this paper, we select the representative LLaMA2-7B-base \cite{touvron2023llama} as the source model to prune, as its size is suitable for experiments and has shown important features of LLMs. We also use the pre-trained OPT-1.3B and OPT-2.7B \cite{zhang2022opt} as the baseline of the pruned models.

We use subsets of RedPajama-V1 \cite{together2023redpajama}, a 1 trillion-token corpus, as the training dataset. Specifically, a subset of 800 million tokens are randomly sampled in the iterative pruning process, while 50 billion tokens are randomly sampled in post-training.
For evaluation, we select held-out downstream tasks from Huggingface open LLM leaderboard\footnote{\ \url{https://huggingface.co/spaces/HuggingFaceH4/open_llm_leaderboard}}, LLaMA2 paper \cite{touvron2023llama}, and Sheared-LLaMA paper \cite{xia2024sheared}. The tasks include zero-shot ARC-E \cite{clark2018think}, BoolQ \cite{clark-etal-2019-boolq}, LogiQA \cite{liu2021logiqa}, OpenbookQA (OBQA) \cite{mihaylov2018can}, PIQA \cite{bisk2020piqa}, SciQ \cite{johannes2017crowd} and few-shot ARC-C \cite{clark2018think}, HellaSwag \cite{zellers2019hellaswag}, TriviaQA \cite{joshi2017triviaqa}, TruthfulQA \cite{lin2022truthfulqa} and WinoGrande \cite{WINOGRANDE}. Details of the evaluation tasks can be found in \autoref{apd:eval}.

\paragraph{Baselines and Implementations}

We compare the following baselines.
(1) LLM-Pruner \cite{ma2023llm}, a structured purning approach with Taylor expansion-based metrics. We reproduce LLM-Pruner with the same architecture as TransAct implementation. (2) Sheared-LLaMA \cite{xia2024sheared}, a masked training-based approach for LLM pruning. We use the open-sourced pruned models and post-train with the same data as TransAct implementation.

The finalized architectures of the pruned models are shown in \autoref{tab:setting} as well as the pruning and training paradigm.
LLM-Pruner and TransAct are implemented in iterative pruning mode, where pruning take place at certain fine-tuning steps. Sheared-LLaMA is reproduced without dynamic batch loading to expose the real performance of pruning without adding influential factors of training. Our implementation is with DeepSpeed on 8 NVIDIA A100 80G GPU, while the sequence length is 4096. Please refer to \autoref{apd:train} for more implementation details\footnote{\ Code available at \url{https://github.com/sbwww/TransAct-pruning}}.

\begin{table*}[ht]
    \renewcommand{\arraystretch}{0.8}
    \setlength\tabcolsep{3pt}
    \centering
    \begin{tabular}{c|cccc|cc|cc}
        \toprule
                                          &
        \multicolumn{6}{c|}{Architecture} &
        \multirow{2}{*}{Pruning}          &
        \multirow{2}{*}{Training}                                                                                                        \\
                                          & $L$ & $H$  & $A$             & $P$   & params. & KV cache
        \\ \midrule
        LLaMA2-7B                         & 32  & 4096 & $32 \times 128$ & 11008 & 6.7B    & 1073M             & -             & -       \\ \midrule
        Sheared-LLaMA-2.7B                & 32  & 2560 & $20 \times 128$ & 6912  & 2.7B    & 671M (-38\%)      & w/ mask       & full FT \\
        LLM-Pruner-2.6B                   & 32  & 4096 & $16 \times 128$ & 3072  & 2.6B    & 536M (\bf{-50\%}) & w/ taylor     & full FT \\
        TransAct-2.6B (ours)              & 32  & 4096 & $16 \times 128$ & 3072  & 2.6B    & 536M (\bf{-50\%}) & w/ activation & full FT \\ \midrule
        Sheared-LLaMA-1.3B                & 24  & 2048 & $16 \times 128$ & 5504  & 1.3B    & 403M (-63\%)      & w/ mask       & full FT \\
        LLM-Pruner-1.3B                   & 32  & 4096 & $6 \times 128$  & 1536  & 1.3B    & 201M (\bf{-81\%}) & w/ taylor     & full FT \\
        TransAct-1.3B (ours)              & 32  & 4096 & $6 \times 128$  & 1536  & 1.3B    & 201M (\bf{-81\%}) & w/ activation & full FT \\ \bottomrule
    \end{tabular}
    \caption{Compressed models with different architectures. $L$ is the number of layers and $H$ is the dimension of hidden states. $A$ denotes the MHA size as $A_n \times A_d$, and the transitional size of MLP is $P$. KV cache is computed with a sequence length of 4096 tokens. B and M stand for billion ($10^9$) and million ($10^6$), respectively.}
    \label{tab:setting}
\end{table*}

\subsection{Experiment Results}

\subsubsection{Efficiency Metrics}

Aiming at efficient deployment of LLMs, the major objective is to reducing inference overheads. In this section, efficiency metrics of FLOPs and end-to-end (E2E) latency of original and pruned models are reported.

\paragraph{FLOPs Reduction}

\begin{figure}[ht]
    \centering
    \includegraphics[width=0.85\columnwidth]{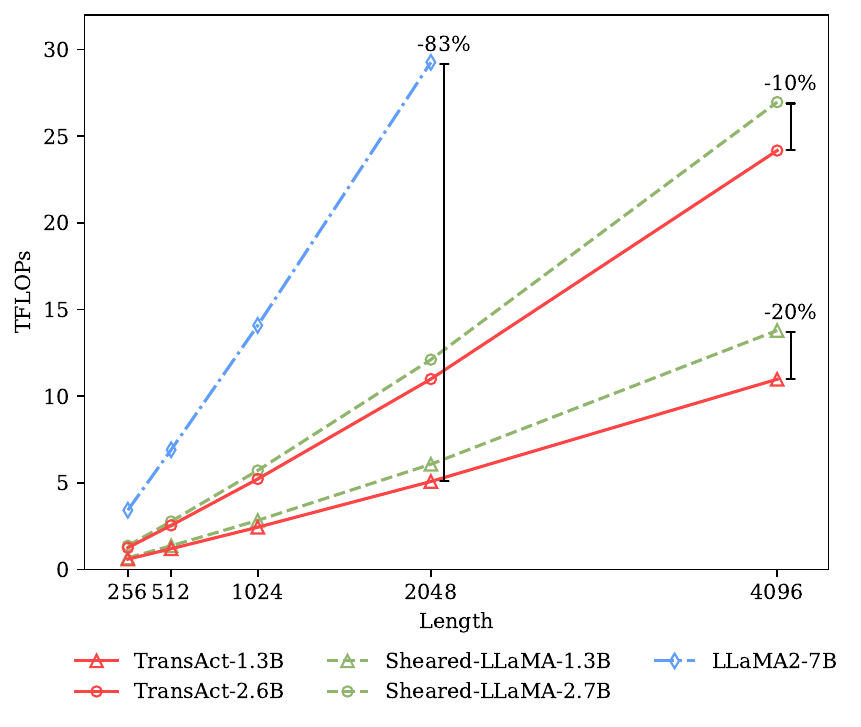}
    \caption{Inference FLOPs of the original and pruned models with variable context length. LLM-Pruner is omitted as the implemented architecture is the same as TransAct counterparts. LLaMA2-7B with 4K-token context is omitted.}
    \label{fig:flops}
\end{figure}

To verify the therotical computation reduction, we profiled the pruned models using PyTorch profiler. Specifically, the profiling is conducted on a single model forward with input length of \{256, 512, 1024, 2048, 4096\} tokens and a single output token. \autoref{fig:flops} elaborates the FLOPs saving of pruned models, where TransAct-1.3B achieves -83\% FLOPs of the original LLaMA model and achieves 20\% addtional FLOPs saving compared to Sheared-LLaMA with similar parameter size. Also, as the context length increases, the FLOPs growth of TransAct remains more steady compared to Sheared-LLaMA. The gradual increase in computation against context length is essential for prevailing LLM applications such as retrieval-augmented generation (RAG) and agent, where the context length commonly exceed 4K, even if the user queries are not necessarily long.

\paragraph{End-side E2E Latency}

\begin{figure*}[ht]
    \centering
    \includegraphics[width=0.7\textwidth]{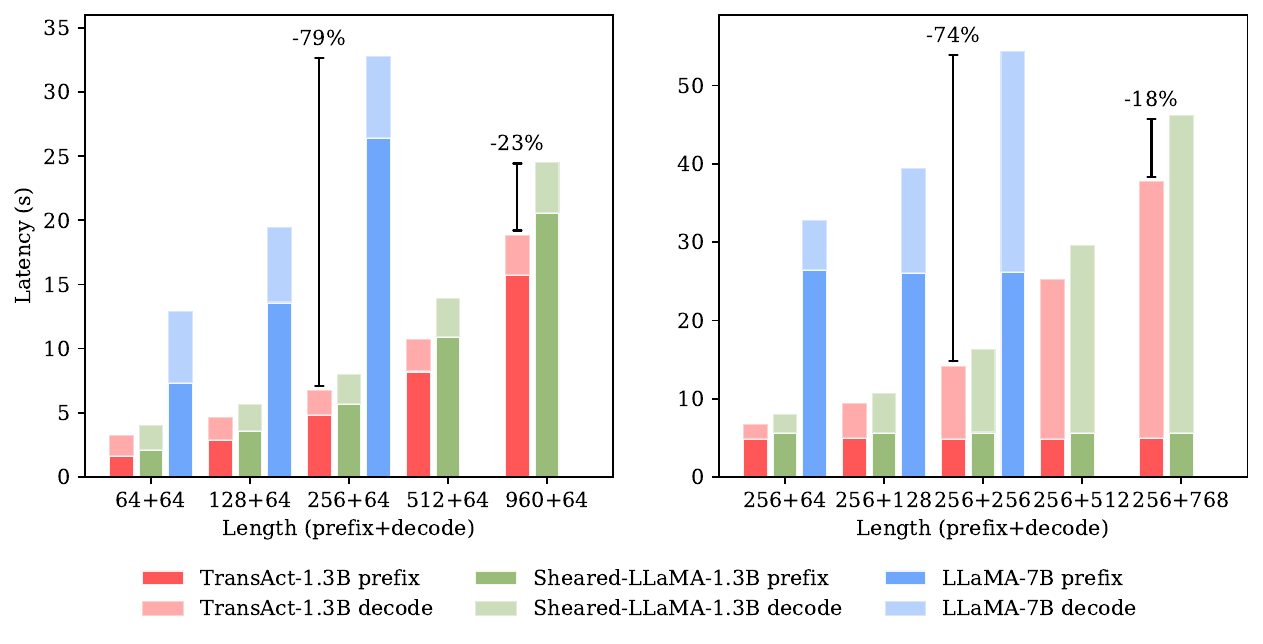}
    \caption{Edge device E2E generation latency (seconds) tested on a Xiaomi 14 mobile phone. The models are quantized to W4A16 by MLC-LLM and tested with variable context length (L, prefix+decode). LLM-Pruner is omitted as the implemented architecture is the same as TransAct counterparts.}
    \label{fig:latency}
\end{figure*}

Beyond therotical analysis, we deploy the models and test the end-to-end inference latency of models. In \autoref{fig:latency}, TransAct-1.3B, Sheared-LLaMA-1.3B and LLaMA2-7B are quantized to W4A16 and deployed on a Xiaomi 14 mobile phone using MLC-LLM \cite{mlc-llm}. TransAct-1.3B has 75\%-80\% time saving compared to the original LLaMA2-7B and 15\%-25\% time saving compared to Sheared-LLaMA, which is consistent to the therotical FLOPs saving and reduced size of model and KV cache.
Compressing the MHA module and KV cache size is crucial on resource-constrained end-side devices as well as the server side. This importance stems from the prevailing deployment approach, which prioritizes weight-only quantization over activation quantization. Weight-only quantization (e.g., W4A16) offers reduced degradation and requires smaller storage compared to activation quantization (e.g., W8A8), making it more favorable for potential mobile applications. Hence, despite the small number of the KV cache compared to model weights, the memory footprint of the 16-bit KV cache is comparable to that of the 4-bit model weights with 4 times amplified.

\paragraph{Server-side E2E Latency}

We deploy TransAct-2.6B, Sheared-LLaMA-2.7B and LLaMA2-7B on a single NVIDIA A100 GPU. The models are tested in the original \texttt{bfloat16} precision. We report the end-to-end generation latency with the batch size of \{4, 8, 16\}, context length of \{1024, 2048\} tokens, and generation length of 2048 tokens.
As shown in \autoref{tab:a100_speed}, TransAct continually outperforms Sheared-LLaMA in variable batch size and context length.

Notably,
an NVIDIA A100 GPU has high bandwidth memory (HBM) of ~2TB/s, which makes the acceleration of TransAct architecture not obvious in streaming mode (i.e., B=1). However,
the inference bottleneck switch from memory bandwidth to computation with larger batch size and context length in server-side applications, and TransAct largely benefits from the reduced computation in the MHA module.

\begin{table*}[ht]

    \centering
    \begin{tabular}{c|ccc|ccc}
        \toprule
                                     &
        \multicolumn{3}{c|}{L=1K+2K} &
        \multicolumn{3}{c}{L=2K+2K}                                                                    \\

                                     & B=1      & B=4       & B=16      & B=1      & B=4       & B=16  \\ \midrule
        LLaMA2-7B                    & 56.9     & 67.0      & 177.2     & 57.0     & 84.0      & OOM   \\ \midrule
        \multirow[c]{2}{*}{Sheared-LLaMA-2.7B}
                                     & 56.2     & 57.5      & 112.1     & 56.5     & 59.7      & 156.6 \\
                                     & (-1.2\%) & (-14.2\%) & (-36.7\%) & (-0.1\%) & (-28.9\%) & -     \\
        \multirow[c]{2}{*}{TransAct-2.6B (ours)}
                                     & 55.0     & 57.1      & 95.0      & 55.1     & 58.4      & 129.0 \\
                                     & (-3.3\%) & (-14.8\%) & (-46.4\%) & (-3.3\%) & (-30.5\%) & -     \\ \bottomrule
    \end{tabular}
    \caption{Server-side E2E generation latency (seconds) tested on an NVIDIA A100 GPU. The models are in \texttt{bfloat16} precision and tested with variable batch size (B) and context length (L, prefix+decode). LLM-Pruner is omitted as the implemented architecture is the same as TransAct counterparts.}
    \label{tab:a100_speed}
\end{table*}

\subsubsection{Performance Metrics}

\begin{table*}[ht]
    \renewcommand{\arraystretch}{0.8}
    \setlength\tabcolsep{4pt}
    \centering
    \begin{tabular}{c|cccccc}
        \toprule
                                     & ARC-E     & BoolQ     & LogiQA    & OBQA       & PIQA       & SciQ                        \\ \midrule
        LLaMA2-7B                    & 74.4      & 80.7      & 30.4      & 43.8       & 76.7       & 94.7                        \\ \midrule
        OPT-2.7B                     & 60.8      & 60.4      & 25.7      & 35.2       & 74.5       & 85.9                        \\
        Sheared-LLaMA-2.7B$^\dagger$ & 66.8      & 66.0      & \bf{28.1} & 38.6       & 76.9       & 89.9                        \\
        LLM-Pruner-2.6B$^*$          & \bf{67.0} & 65.9      & 27.7      & \bf{38.8}  & \bf{77.1}  & 90.1                        \\
        TransAct-2.6B (ours)         & 65.5      & \bf{66.3} & 27.9      & 38.2       & 76.9       & \bf{91.0}                   \\ \midrule
        OPT-1.3B                     & 57.1      & 57.7      & 27.0      & 33.4       & 72.4       & 84.4                        \\
        Sheared-LLaMA-1.3B$^\dagger$ & 59.3      & 61.6      & 27.5      & 33.0       & 74.2       & 85.8                        \\
        LLM-Pruner-1.3B$^*$          & \bf{60.0} & 59.5      & \bf{28.7} & \bf{35.2}  & 73.6       & 86.1                        \\
        TransAct-1.3B (ours)         & 57.4      & \bf{63.4} & 27.5      & 33.8       & \bf{74.4}  & \bf{86.7}                   \\ \midrule
        \multirow[c]{2}{*}{Model}
                                     & ARC-C     & HellaSwag & TriviaQA  & TruthfulQA & WinoGrande & \multirow[c]{2}{*}{Average} \\
                                     & (25)      & (10)      & (5)       & $**$       & (5)        &                             \\ \midrule
        LLaMA2-7B                    & 53.4      & 78.6      & 55.1      & 44.6       & 72.3       & 64.1                        \\ \midrule
        OPT-2.7B                     & 34.0      & 61.4      & 23.7      & \bf{37.6}  & 61.7       & 51.0                        \\
        Sheared-LLaMA-2.7B$^\dagger$ & \bf{40.0} & 71.0      & 21.2      & 32.0       & 65.0       & 54.1                        \\
        LLM-Pruner-2.6B$^*$          & 38.6      & 70.8      & 17.3      & 32.9       & 63.6       & 53.6                        \\
        TransAct-2.6B (ours)         & 38.9      & \bf{71.2} & \bf{33.9} & 33.6       & \bf{65.5}  & \bf{55.3}                   \\ \midrule
        OPT-1.3B                     & 29.7      & 54.6      & 16.7      & 38.7       & \bf{60.0}  & 48.3                        \\
        Sheared-LLaMA-1.3B$^\dagger$ & 30.3      & \bf{62.6} & 14.0      & 34.1       & 59.3       & 49.2                        \\
        LLM-Pruner-1.3B$^*$          & 30.3      & 59.0      & 7.9       & 35.9       & 56.4       & 48.4                        \\
        TransAct-1.3B (ours)         & \bf{32.2} & 59.9      & \bf{18.4} & \bf{39.6}  & 56.5       & \bf{50.0}                   \\ \bottomrule
    \end{tabular}
    \caption{Zero-shot and few-shot evaluation results on standard benchmarks. LLaMA2-7B and OPT models are pre-trained models used as the baseline. Results of LLM-Pruner$^*$ are reproduced by us with the same architecture as TransAct, while Sheared-LLaMA$^\dagger$ models are post-trained from released pruned models. (N) below task name indicates N-shots evaluation, TruthfulQA$^{**}$ prepends 6 examples even in the zero-shot setting. The best results are in \textbf{bold}.}
    \label{tab:result}
\end{table*}

The evaluation results of pruned models on held-out benchmarks are listed in \autoref{tab:result} while the perplexity of language modeling tasks are in \autoref{apd:ppl}.
On few-shot tasks, TransAct-2.6B achieves the best performance performance compared to SOTA approaches.
TransAct exhibits a significant leap over LLM-Pruner and Sheared-LLaMA on TriviaQA and TruthfulQA, which evaluate the truthfulness and world knowledge of the LLM. Whereas, the pre-trained OPT models achieve the highest metric on the two tasks although other abilities are inferior to the pruned models. We interpret that TransAct better preserved the world knowledge of the original LLM, which is much harder than preserving language modeling and commonsense reasoning capabilities.
At 80\% compression, TransAct-1.3B achieves 78.0\% performance of LLaMA2-7B on average, addressing the effectiveness of TransAct at highly compressed settings. Whereas LLM-Pruner fails at most few-shot tasks. Thereby, we address the inapplicability of structured pruning with the Taylor expansion-based metric. LLMs are fundamentally pre-trained on a large corpus to obtain world knowledge. However, the Taylor expansion-based metric, which guides the pruning by minimizing the approximated language modeling loss on a small calibration set, fails to preserve knowledge and degrade the pruned LLM. Amplifying the calibration set by a significant order of magnitude is an intuitive solution. However, the computation of Jacobian and Hessian matrices of LLM weights on a large calibration set is enormous.

Notably, the reproduced LLM-Pruner-2.6B with iterative pruning reaches 83.6\% performance of the uncompressed LLaMA2-7B. Whereas in its original paper, the performance at 50\% compression ratio can barely reach 78\% of the original model \cite{ma2023llm}. The results strengthen the necessity of iterative pruning at LLM structured pruning. Specifically, iterative pruning is gradual and conservative at each step, lessening the approximation error of pruning metrics.

\begin{figure}[ht]
    \centering
    \includegraphics[width=0.82\columnwidth]{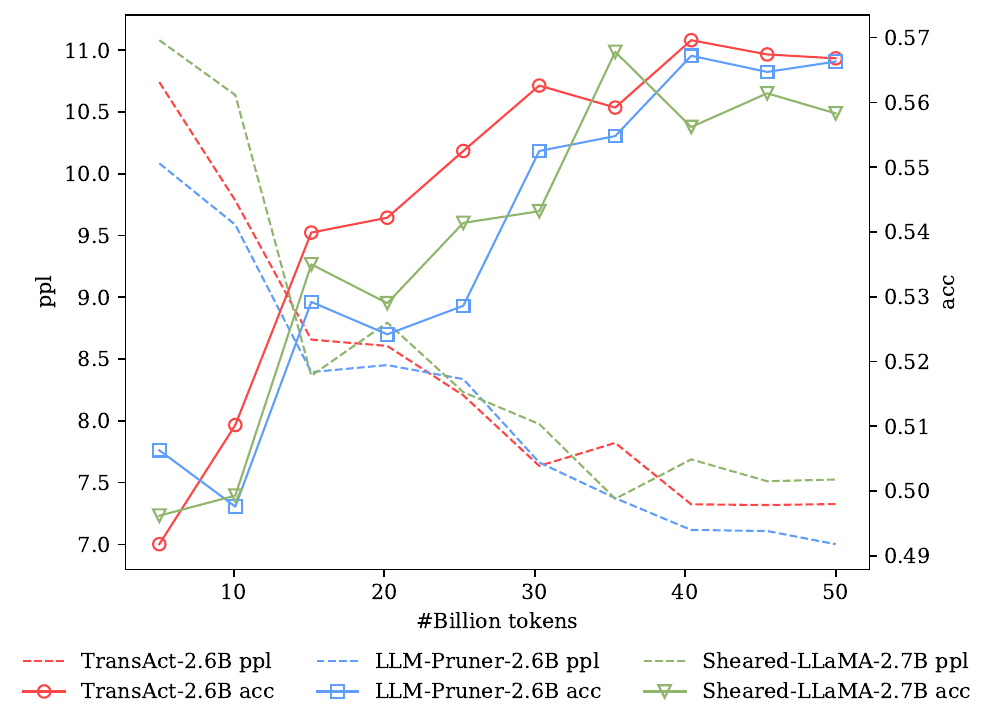}
    \caption{LAMBADA perplexity and accuracy on every checkpoint of TransAct-2.6B, LLM-Pruner-2.6B and Sheared-LLaMA-2.7B post-training.}
    \label{fig:curve}
\end{figure}

\autoref{fig:curve} illustrate the zero-shot LAMBADA language modeling performance at each checkpoint of the pruned model post-training.
Although TransAct-2.6B has a clear advantage between 10b to 30b tokens trained, the gap between different pruning approaches diminishes as the pruned model is gradually recovered by post-training. Notably, the result of LLM-Pruner-2.6B exhibits the lowest perplexity in \autoref{fig:curve}. However, it does not necessarily indicate the highest accuracy on LAMBADA, nor the performance on other tasks.

\subsubsection{Ablation Studies}

We conduct a comprehensive evaluation of the pruned LLM, considering factors of pruning shots, calibration samples, and the pruning ratio of each module. The findings provide insights for the further development of compact LMs.

\paragraph{Impact of Iterative Pruning}

While LLM-Pruner has demonstrated a close performance gap to the original model at a moderate ratio of 20\%, the significant performance degradation observed at over 50\% pruned is far from acceptable in the original implementation \cite{ma2023llm}. However, the results in \autoref{tab:result} indicate that LLM-Pruner achieves comparable performance to the SOTA approach Sheared-LLaMA even at a compression ratio of 85\%. This achievement can be attributed to our iterative implementation of pruning.

To further verify the effectiveness of iterative pruning, we conduct experiments on LLM-Pruner-2.6B and TransAct-2.6B with different numbers of pruning shots. Specifically, we explore pruning shots ranging from $\{1, 2, 4, 8, 16\}$. Except for single-shot pruning, all others have a total of 800 million tokens throughout the iterative pruning stage. After the final pruning, all models undergo full fine-tuning with 200 million tokens.

Sheared-LLaMA is considered an $\infty$-shot pruning approach with all the parameters trained and is not compared.

\begin{figure}[ht]
    \centering
    \includegraphics[width=0.99\columnwidth]{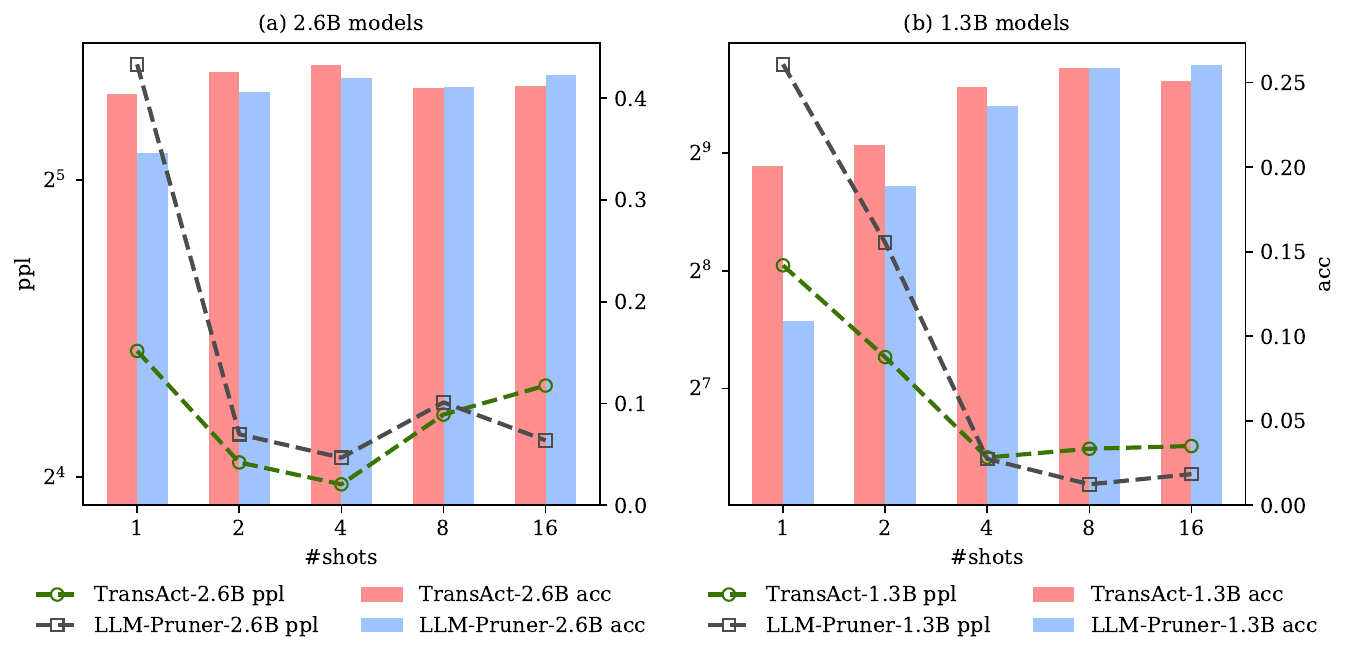}
    \caption{LAMBADA perplexity and accuracy on models with different numbers of pruning shots.}
    \label{fig:abl_shots}
\end{figure}

Results in \autoref{fig:abl_shots} indicate the relationship between pruning shots and performance on LAMBADA language modeling. Although iterative pruning is beneficial, the pruning shots need to be controlled with a total number of tokens is fixed. The performance of 2.6B models degrades when the pruning shot is increased from 4 to 8. The rationale of this phenomenon is that when training is insufficient between two pruning shots, the pruning would be misguided and the pruned model would exhibit a degradation. Whereas, for 1.3B models, the performance exhibits a slight degradation at 16 shots, indicating the benefit of increased shots has not yet been overwhelmed by the insufficiency of training data. LLM-Pruner has a slight advantage over TransAct at 16 shots pruning, as fewer parameters pruned at each shot reduce the approximation error of loss with Taylor expansion.

\paragraph{Impact of Calibration Samples}

To evaluate the sensitivity of pruning approaches to calibration samples used in the pruning process, we conduct single-shot pruning experiments on different numbers of calibration samples. 200 million tokens are used for the restoration after pruning.

\begin{figure}[ht]
    \centering
    \includegraphics[width=0.99\columnwidth]{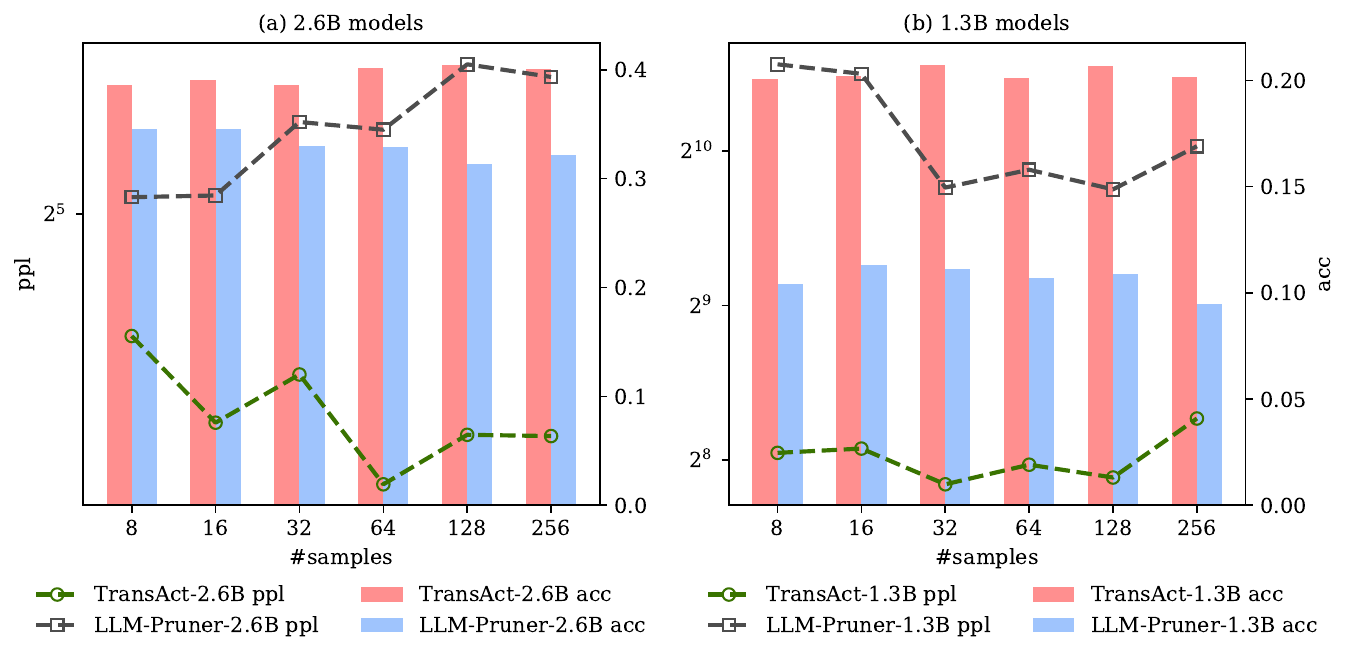}
    \caption{LAMBADA perplexity and accuracy on models with different numbers of calibration samples.}
    \label{fig:abl_sample}
\end{figure}

The results in \autoref{fig:abl_sample} indicate that increasing the sample size can bring gains, but the marginal benefits decrease after increasing to 128 samples. When leveraging 256 samples, the performance of both TransAct and LLM-Pruner degrade. Also, the degradation trend is more obvious on LLM-Pruner than on TransAct. We attribute this to early overfitting of calibration samples, where the pruning guided by Taylor expansion of loss quickly overfits on the calibration set, and the calibration samples are not large enough to exhibit diversity. As pruning is efficient in our implementation, we prefer using 128 samples for the pruning metric, which can be computed in less than 1 minute on a single A100 GPU to prune LLaMA2-7B.

\paragraph{Analysis on Module Redundancy}

\begin{figure}[ht]
    \centering
    \includegraphics[width=0.75\columnwidth]{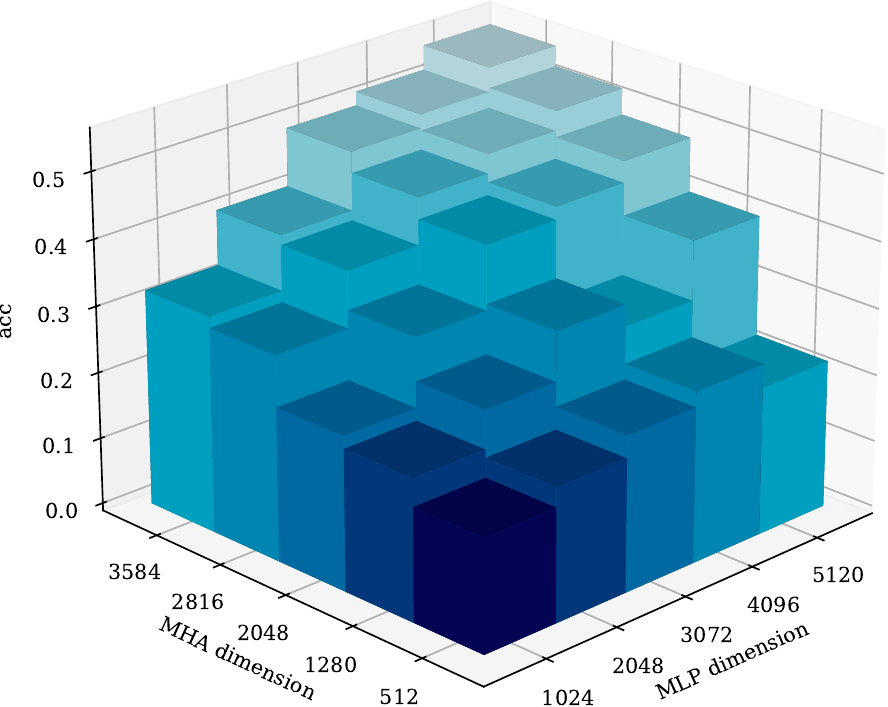}
    \caption{LAMBADA accuracy on 25 pruned models with different architectures. Bars with the same color indicate models with the same number of parameters.}
    \label{fig:abl_module}
\end{figure}

To validate the redundancy of pre-trained models and help future compact model design, we conduct experiments on different compression ratios of MHA and MLP modules. Specifically, after single-shot pruning with TransAct and post-training on 200 million tokens, the accuracy of LAMBADA language modeling is evaluated. Specifically, using the shape of our TransAct-2.6B as the center point, we vary the MHA dimension $A$ ranged from $\{512, 1280, 2048, 2816, 3584\}$ with the head dimension of 128. And, the MLP dimension $P$ is set to $\{1024, 2048, 3072, 4096, 5120\}$. These configurations resulted in 25 distinct models obtained by pairwise combinations. Notably, the 25 models are organized into 9 groups, each containing an e number of parameters. These groups are visually distinguished by color in \autoref{fig:abl_module}.

The results presented in \autoref{fig:abl_module} reveal a clear trend that, the models at the center exhibit the best performance within each group, and in some cases, even surpass models of larger sizes. For instance, the combination of 2048A-3072P (i.e., TransAct-2.6B) model surpasses both 3584A-2048P and 1280A-5120P (2.9B) models. Also, when pruning the MHA intermediate size to 512, the performance drops to the worst within each group. We interpret that MHA functions as the crucial module of Transformer-based LLMs while MLP has a larger redundancy that can be compressed. Further, the findings indicate that models with a uniform MHA and MLP size generally outperform the others. For 2048A-3072P, an MHA module has 33.5 million parameters and an MLP module has 37.7 million parameters. On the contrary, extreme pruning of either MHA or MLP alone leads to severe performance degradation. Hence, the collaborative compression of both MHA and MLP is encouraged.

\section{Conclusion}

In this paper, we introduce TransAct, an effective and efficient pruning approach coupled with an architecture designed for pruned LLMs. TransAct compresses the original LLM into a compact dense model with an intra-module low-rank architecture, achieving the fastest inference speed and lowest overheads compared to models of similar sizes. The compression is guided by the magnitudes of the transitional activations within the MHA and MLP modules. Specifically, intra-module dimensions with small activations are structurally pruned out, while inter-module dimensions are preserved.
Experiments on open-source LLMs and downstream benchmarks demonstrate the strength of our approach, particularly at high compression rates. Also, we thoroughly evaluated the pruned LLM with respect to calibration samples, pruning ratio, and pruning shots. The results provide insights and experimental results for further development of compact yet powerful LMs.

\section*{Limitations}

Although TransAct is found effective in the experiments, some points are not fully covered in this paper. We list the limitations and future directions as follows.
(1) TransAct is a static pruning approach where the computation of the pruned LLM is irrelvant to input instances. However, recent research progress in MoE \cite{jiang2024mixtral} indicates that dynamically compressed models are model powerful than statically compressed ones. Hence, a pruning approach integrating static and dynamic compression with approporate ratio can be further studied.
(2) TransAct is targeted to Transformer-based LLMs. Different architectures including RWKV \cite{peng2023rwkv}, Mamba \cite{gu2023mamba} are not yet investigated.
(3) The pruning of TransAct is conducted on base models. Structurally pruning a human-aligned LLM still remains challenging, mainly because of the inconsistency in training data paradigm between pre-training and alignment.

\bibliography{anthology,custom}

\vfill\break

\appendix

\section{Details of Training Arguments}\label{apd:train}

The training arguments are listed in \autoref{tab:train}. The experiments are conducted on Huggingface Transformers with DeepSpeed and FlashAttention2 integration. We set the training arguments based on accessible computational resources and setting of \citet{xia2024sheared}. There is no hyperparameters searching or tuning in this work, and we believe it is potentially beneficial to tune the hyperparameters with sufficient resources.

\begin{table}[ht]
    \setlength\tabcolsep{5pt}
    \centering
    \begin{tabular}{c|c}
        \toprule
        Argument                 & Value             \\ \midrule
        Length                   & 4096              \\
        N GPUs                   & 8                 \\
        Global batch size        & 64                \\
        Optimizer                & AdamW             \\
        $\beta_1, \beta_2$       & 0.9, 0.95         \\
        Learning rate            & 5e-5              \\
        Learning rate schelduler & Cosine            \\
        Warmup                   & 0.03              \\
        Data type                & \texttt{bfloat16} \\
        DeepSpeed                & Zero-2            \\
        Attention implementation & FlashAttention2   \\
        \bottomrule
    \end{tabular}
    \caption{Details of training arguments.}
    \label{tab:train}
\end{table}

\section{Details of Evaluation Tasks}\label{apd:eval}

The downstream tasks used for evaluation are listed in \autoref{tab:dataset}. The evaluations are conducted based on lm-evaluation-harness \footnote{\url{https://github.com/EleutherAI/lm-evaluation-harness}} repository with MIT license. In \autoref{tab:dataset}, "acc\_norm" stands for accuracy after normalization by byte-length, "em" stands for exact match, and "mc2" stands for the normalized probability assigned to all true answers in multiple choices \cite{lin2022truthfulqa}.

\begin{table*}[ht]
    \centering
    \begin{tabular}{c|ccc|ccc}
        \toprule
        \multirow[c]{2}{*}{Task}              &
        \multicolumn{3}{c|}{Used by}          &
        \multirow[c]{2}{*}{\#samples}         &
        \multirow[c]{2}{*}{\#shots}           &
        \multirow[c]{2}{*}{Metric}                                                                                   \\
                                              & (1)          & (2)          & (3)          &       &                 \\ \midrule

        ARC-C \cite{clark2018think}           & $\checkmark$ & $\checkmark$ & $\checkmark$ & 1172  & 25 & acc\_norm  \\
        ARC-E \cite{clark2018think}           &              & $\checkmark$ & $\checkmark$ & 2376  & -  & acc        \\
        BoolQ \cite{clark-etal-2019-boolq}    &              & $\checkmark$ & $\checkmark$ & 3270  & -  & acc        \\
        HellaSwag \cite{zellers2019hellaswag} & $\checkmark$ & $\checkmark$ & $\checkmark$ & 10042 & 10 & acc\_norm  \\
        LAMBADA \cite{paperno2016lambada}     &              &              & $\checkmark$ & 5153  & -  & ppl \& acc \\
        LogiQA \cite{liu2021logiqa}           &              & $\checkmark$ & $\checkmark$ & 651   & -  & acc\_norm  \\
        OBQA \cite{mihaylov2018can}           &              & $\checkmark$ &              & 500   & -  & acc\_norm  \\
        PIQA \cite{bisk2020piqa}              &              & $\checkmark$ & $\checkmark$ & 1838  & -  & acc\_norm  \\
        SciQ \cite{johannes2017crowd}         &              &              & $\checkmark$ & 1000  & -  & acc        \\
        TriviaQA \cite{joshi2017triviaqa}     &              & $\checkmark$ &              & 11313 & 5  & em         \\
        TruthfulQA \cite{lin2022truthfulqa}   & $\checkmark$ & $\checkmark$ &              & 817   & *  & mc2        \\
        WikiText \cite{merity2016pointer}     &              &              &              & 62    & -  & ppl        \\
        WinoGrande \cite{WINOGRANDE}          & $\checkmark$ & $\checkmark$ & $\checkmark$ & 1267  & 5  & acc        \\
        \bottomrule
    \end{tabular}
    \caption{Details of evaluation tasks. (1), (2) and (3) refer to Open LLM Leaderboard, LLaMA2 paper and Sheared-LLaMA paper, respectively. * TruthfulQA prepends 6 examples even in zero-shot setting.}
    \label{tab:dataset}
\end{table*}

\section{Perplexity of Language Modeling}\label{apd:ppl}

To evaluate the basic ability of language modeling, we test perplexity of models on WikiText and LAMBADA corpus, and the results are in \autoref{tab:result_ppl}. WikiText contains long documents that exceed the maximum length of LLaMA2 (i.e., 4K), and the documents are truncated into three set of maximum length, \{1K, 2K, 4K\}. Samples in LAMBADA test set are below 1K tokens, so the above three set of maximum length does not effect the results.

\begin{table*}[ht]
    \centering
    \begin{tabular}{c|ccc|c}
        \toprule
                                     & WikiText-1K & WikiText-2K & WikiText-4K & LAMBADA   \\ \midrule
        LLaMA2-7B                    & 13.8        & 12.3        & 11.7        & 4.0       \\ \midrule
        Sheared-LLaMA-2.7B$^\dagger$ & 16.3        & 14.7        & 13.9        & 7.5       \\
        LLM-Pruner-2.6B$^*$          & 16.0        & 15.1        & 14.2        & \bf{7.1}  \\
        TransAct-2.6B (ours)         & \bf{15.3}   & \bf{13.8}   & \bf{13.1}   & 7.4       \\ \midrule
        Sheared-LLaMA-1.3B$^\dagger$ & 20.0        & 18.1        & 17.1        & \bf{11.4} \\
        LLM-Pruner-1.3B$^*$          & 21.7        & 20.5        & 19.8        & 13.2      \\
        TransAct-1.3B (ours)         & \bf{19.3}   & \bf{17.3}   & \bf{16.3}   & 13.0      \\ \bottomrule
    \end{tabular}
    \caption{Perplexity evaluation results on standard corpus. LLaMA2-7B is a pre-trained model used as the baseline. Results of LLM-Pruner$^*$ are reproduced by us with the same architecture as TransAct, while Sheared-LLaMA$^\dagger$ models are post-trained from released pruned models. Suffix -NK indicates the maximum length of test samples, LAMBADA test set samples are below 1K tokens. The best results are in \textbf{bold}.}
    \label{tab:result_ppl}
\end{table*}


\end{document}